\documentclass[11pt,a4paper]{article}
\usepackage[hyperref]{acl2019}
\usepackage{times}
\usepackage{latexsym}
\usepackage{soul}
\usepackage{booktabs}
\usepackage{graphicx,graphics,wrapfig,epstopdf,subfig}
\usepackage{algorithm,algpseudocode}
\usepackage{url}

\newcommand{\beq}{\begin{equation}}
\newcommand{\eeq}{\end{equation}}
\newcommand{\beqs}{\begin{eqnarray}}
\newcommand{\eeqs}{\end{eqnarray}}
\newcommand{\barr}{\begin{array}}
	\newcommand{\earr}{\end{array}}

\newcommand{\eg}{\textit{e.g.}}
\newcommand{\ie}{\textit{i.e.}}

\usepackage{amsmath,amsfonts,bm}

\def\eqref#1{equation~\ref{#1}}

\def\1{\bm{1}}

\DeclareMathAlphabet{\mathsfit}{\encodingdefault}{\sfdefault}{m}{sl}
\SetMathAlphabet{\mathsfit}{bold}{\encodingdefault}{\sfdefault}{bx}{n}

\newcommand{\Amat}{{\bf A}}
\newcommand{\Bmat}{{\bf B}}
\newcommand{\Cmat}{{\bf C}}

\newcommand{\Hmat}{{\bf H}}

\newcommand{\Qmat}{{\bf Q}}

\newcommand{\Smat}{{\bf S}}
\newcommand{\Tmat}{{\bf T}}

\newcommand{\Wmat}{{\bf W}}

\newcommand{\Zmat}{{\bf Z}}

\newcommand{\hv}{{\boldsymbol h}}

\newcommand{\sv}{{\boldsymbol s}}

\newcommand{\wv}{{\boldsymbol w}}

\newcommand{\xv}{{\boldsymbol x}}
\newcommand{\yv}{{\boldsymbol y}}

\newcommand{\zv}{{\boldsymbol z}}

\newcommand{\muv}{{\boldsymbol \mu}}
\newcommand{\nuv}{{\boldsymbol \nu}}

\aclfinalcopy 
\def\aclpaperid{1837} 

\usepackage{amssymb}
\newcommand{\CG}{\mathcal{G}}
\newcommand{\CV}{\mathcal{V}}
\newcommand{\CE}{\mathcal{E}}
\newcommand{\CT}{\mathcal{T}}
\newcommand{\BR}{\mathbb{R}}
\newcommand{\CA}{\mathcal{A}}
\newcommand{\EE}{\mathbb{E}}
\newcommand{\CX}{\mathcal{X}}
\newcommand{\CY}{\mathcal{Y}}

\newcommand{\Enc}{\text{Enc}}
\newcommand{\raw}{\text{raw}}

\title{Improving Textual Network Embedding with Global Attention\\via Optimal Transport}

\author{Liqun Chen, Guoyin Wang, Chenyang Tao, Dinghan Shen, \\ \textbf{Pengyu Cheng}, \textbf{Xinyuan Zhang}, \textbf{Wenlin Wang}, \textbf{Yizhe Zhang}, \textbf{Lawrence Carin}\\
\texttt{\{liqun.chen\}@duke.edu} \\
}

\date{}

\begin{document}
\maketitle
\begin{abstract}
Constituting highly informative network embeddings is an important tool for network analysis. It encodes network topology, along with other useful side information, into low-dimensional node-based feature representations that can be exploited by statistical modeling. This work focuses on learning context-aware network embeddings augmented with text data. We reformulate the network-embedding problem, and present two novel strategies to improve over traditional attention mechanisms: ($i$) a content-aware sparse attention module based on optimal transport, and ($ii$) a high-level attention parsing module. Our approach yields naturally sparse and self-normalized relational inference. It can capture long-term interactions between sequences, thus addressing the challenges faced by existing textual network embedding schemes.
Extensive experiments are conducted to demonstrate our model can consistently outperform alternative state-of-the-art methods.

\end{abstract}
\vspace{-2mm}
\section{Introduction}
\vspace{-2mm}
When performing network embedding, one maps network nodes into vector representations that reside in a low-dimensional latent space. Such techniques seek to encode topological information of the network into the embedding, such as affinity \citep{tang2009relational}, local interactions ({\it e.g,} local neighborhoods) \citep{perozzi2014deepwalk}, and high-level properties such as community structure \citep{wang2017community}. Relative to classical network-representation learning schemes \citep{zhang2018network}, network embeddings provide a more fine-grained representation that can be easily repurposed for other downstream applications ({\it e.g.,} node classification, link prediction, content recommendation and anomaly detection). 

For real-world networks, one naturally may have access to rich side information about each node. Of particular interest are {\it textual networks}, where the side information comes in the form of natural language sequences \citep{le2014probabilistic}. For example, user profiles or their online posts on social networks ({\it e.g.,} Facebook, Twitter), and documents in citation networks ({\it e.g.,} Cora, arXiv). The integration of text information promises to significantly improve embeddings derived solely from the noisy, sparse edge representations \citep{yang2015network}. 

Recent work has started to explore the joint embedding of network nodes and the associated text for abstracting more informative representations. \citet{yang2015network} reformulated DeepWalk embedding as a matrix factorization problem, and fused text-embedding into the solution, while \citet{sun2016general} augmented the network with documents as auxiliary nodes. Apart from direct embedding of the text content, one can first model the topics of the associated text \citep{blei2003latent} and then supply the predicted labels to facilitate embedding \citep{tu2016max}. 

Many important downstream applications of network embeddings are {\it context-dependent}, since a static vector representation of the nodes adapts to the changing context less effectively \citep{tu2017cane}. For example, the interactions between social network users are context-dependent ({\it e.g.,} family, work, interests), and contextualized user profiling can promote the specificity of recommendation systems. This motivates context-aware embedding techniques, such as CANE \citep{tu2017cane}, where the vector embedding dynamically depends on the context. For textual networks, the associated texts are natural candidates for context. CANE introduced a simple mutual attention weighting mechanism to derive context-aware dynamic embeddings for link prediction. Following the CANE setup, WANE \citep{shen2018improved} further improved the contextualized embedding, by considering fine-grained text alignment. 

Despite the promising results reported thus far, we identify three major limitations of existing context-aware network embedding solutions. First, mutual (or cross) attentions are computed from pairwise similarities between local text embeddings (word/phrase matching), whereas global sequence-level modeling is known to be more favorable across a wide range of NLP tasks ~\citep{maccartney2009natural, liu2018stochastic, malakasiotis2007learning,guo2018soft}. 
Second, related to the above point, low-level affinity scores are directly used as mutual attention without considering any high-level parsing. Such an over-simplified operation denies desirable features, such as noise suppression and relational inference \citep{santoro2017simple}, thereby compromising model performance. 
Third, mutual attention based on common similarity measures (\eg, cosine similarity) typically yields dense attention matrices,
while psychological and computational evidence suggests a sparse attention mechanism functions more effectively~\citep{martins2016softmax, niculae2017regularized}. 
Thus such naive similarity-based approaches can be suboptimal, since they are more likely to incorporate irrelevant word/phrase matching.
\vspace{-1mm}

This work represents an attempt to improve context-aware textual network embedding, by addressing the above issues. Our contributions include: ($i$) We present a principled and more-general formulation of the network embedding problem, under reproducing kernel Hilbert spaces (RKHS) learning; this formulation clarifies aspects of the existing literature and provides a flexible framework for future extensions. ($ii$) A novel global sequence-level matching scheme is proposed, based on optimal transport, which matches key concepts between text sequences in a sparse attentive manner. ($iii$) We develop a high-level attention-parsing mechanism that operates on top of low-level attention, which is capable of capturing long-term interactions and allows relational inference for better contextualization. We term our model {\it \textbf{G}lobal \textbf{A}ttention \textbf{N}etwork \textbf{E}mbedding}  (\textbf{GANE}).
To validate the effectiveness of GANE, we benchmarked our models against state-of-the-art counterparts on multiple datasets. Our models consistently outperform competing methods. 

\section{Problem setup}
\vspace{-2mm}
\label{sec:problem_setup}
We introduce basic notation and definitions used in this work. 
\vspace{-2mm}
\paragraph{Textual networks.} Let $\CG = (\CV, \CE, \CT)$ be our textual network, where $\CV$ is the set of nodes, $\CE \subseteq \CV \times \CV$ are the edges between the nodes, and $\CT = \{ S_v \}_{v\in\CV}$ are the text data associated with each node. We use $S_v = [\omega_1, \cdots, \omega_{n_v}]$ to denote the token sequence associated with node $v \in \CV$, of length $n_v = |S_v|$ where $| \cdot |$ denotes the counting measure. To simplify subsequent discussion, we assume all tokens have been pre-embedded in a $p$-dimensional feature space. As such, $S_v$ can be directly regarded as a $\BR^{p \times n_v}$ matrix tensor. We use $\{ u, v\}$ to index the nodes throughout the paper. We consider directed unsigned graphs, meaning that for each edge pair $(u,v) \in \CE$ there is a nonnegative weight $w_{uv}$ associated with it, and $w_{uv}$ does not necessarily equal $w_{vu}$. 
\vspace{-2mm}
\paragraph{Textual network embedding.} The goal of textual network embedding is to identify a $d$-dimensional embedding vector $\zv_v \in \BR^d$ for each node $v \in \CV$, which encodes network topology ($\CE$) via leveraging information from the associated text ($\CT$). In mathematical terms, we want to learn an encoding (embedding) scheme $\Zmat_{\CG} \triangleq \{ \zv_v = \Enc(v;\CG) \}_{v\in \CV}$ and a probabilistic decoding model with likelihood $p_{\theta}(E;\Zmat)$, where $E \subseteq \CV \times \CV$ is a random network topology for node set $\CV$, such that the likelihood for the observed topology $p_{\theta}(\CE|\Zmat_{\CG})$ is high. Note that for efficient coding schemes, the embedding dimension is much smaller than the network size (\ie, $d \ll | \CV |$). In a more general setup, the decoding objective can be replaced with $p_{\theta}(\CA|\Zmat)$, where $\CA$ denotes observed attributes of interest (\eg, node label, community structure, {\it etc.}). 

\vspace{-2mm}
\paragraph{Context-aware embedding.} One way to promote coding efficiency is to contextualize the embeddings. More specifically, the embeddings additionally depend on an exogenous context $c$. To distinguish it from the context-free embedding $\zv_u$, we denote the context-aware embedding as $\zv_{u|c}$, where $c$ is the context. For textual networks, when the embedding objective is network topology reconstruction, a natural choice is to treat the text as context \citep{tu2017cane}. In particular, when modeling the edge $w_{uv}$, $S_v$ and $S_u$ are respectively treated as the context for context-aware embeddings $\zv_{u|c}$ and $\zv_{v|c}$, which are then used in the prediction of edge likelihood. 

\vspace{-2mm}
\paragraph{Attention \& text alignment.} Much content can be contained in a single text sequence, and retrieving them with a fixed length feature vector can be challenging. A more flexible solution is to employ an attention mechanism, which only attends to content that is relevant to a specific query ~\cite{vaswani2017attention}. Specifically, attention models leverage a gating mechanism to de-emphasize irrelevant parts in the input; this method pools information only from the useful text, which is also a fixed length vector but that only encodes information with respect to one specific content ~\cite{santos2016attentive}. Popular choices of attention include normalized similarities in the feature space ({\it e.g.}, Softmax normalized cosine distances). For two text sequences, one can build a mutual attention by cross-relating the content from the respective text ~\citep{santoro2017simple}. In text alignment, one further represents the content from one text sequence using the mutual attention based attentive-pooling on the other sequence \citep{shen2018improved}. 

\vspace{-2mm}
\paragraph{Optimal transport (OT).} 
Consider $\muv = \{(\xv_i, \mu_i)\}_{i=1}^n$ and $\nuv = \{ (\yv_j, \nu_j) \}_{j=1}^m$, a set of locations and their associated nonnegative mass (we assume $\sum_i \mu_i = \sum_j \nu_j = 1$). We call $\pi \in \BR_+^{n \times m}$ a valid transport plan if it properly redistributes mass from $\muv$ to $\nuv$, \ie, $\sum_i \pi_{ij} = \nu_j$ and $\sum_j \pi_{ij} = \mu_i$. In other words, $\pi$ breaks mass at $\{\xv_i\}$ into smaller parts and transports $\pi_{ij}$ units of $\xv_i$ to $\yv_j$. Given a cost function $c(\xv,\yv)$ for transporting unit mass from $\xv$ to $\yv$, discretized OT solves the following constrained optimization for an optimal transport plan $\pi^*$~\citep{peyre2017computational}:  
\begin{equation}
\label{eq:ot}
    \mathcal{D}_c(\muv,\nuv) = \inf_{\pi\in\Pi(\mu,\nu)}\left\{ \sum_{ij} \pi_{ij} c(\xv_i,\yv_j)\right\} \,,
\end{equation} 
where $\Pi(\muv,\nuv)$ denotes the set of all viable transport plans. Note that $c(\xv,\yv)$ is a distance metric on $\CX$, and $D_c(\muv,\nuv)$ induces a distance metric on the space of probability distributions supported on $\CX$, commonly known as the Wasserstein distance~\citep{villani2008optimal}. Popular choices of cost include Euclidean cost $\| \xv - \yv \|_2^2$ for general probabilistic learning \citep{gulrajani2017improved} and cosine similarity cost $\cos(\xv,\yv)$ for natural language models \citep{chen2018adversarial}. Computationally, OT plans are often approximated with Sinkhorn-type iterative schemes \citep{cuturi2013sinkhorn}. Algorithm \ref{alg:ipot} summarizes a particular variant used in our study \citep{xie2018fast}. 

\begin{algorithm}[!t]
\footnotesize
\caption{\footnotesize Optimal transport solver (SolveOT)}
\label{alg:ipot}
\begin{algorithmic}[1]
\State {\bfseries Input:} \footnotesize{\footnotesize{Sentence matrices 
 $\Smat=\{\wv_i\}_1^n$, $\Smat'=\{\wv'_j\}_1^m$ $\,\,\,\,$ and generalized stepsize $1/\beta$, 
}}

\State $\boldsymbol{\sigma}=\frac{1}{m}\mathbf{1_m}$, $\Tmat^{(1)} = \mathbf{1_n} \mathbf{1_m}^\top$
\State $\Cmat_{ij} = c(\zv_i, \zv'_j)$, $\Amat_{ij} = {\rm e}^{-\frac{\Cmat_{ij}}{\beta}}$
\For{$t=1,2,3\ldots$}
    \State $\Qmat = \Amat \odot \Tmat^{(t)}$ \footnotesize{// $\odot$ is Hadamard product}
    \For{$k=1,\ldots K$} \footnotesize{// $K=1$ in practice}
        \State $\boldsymbol{\delta} = \frac{1}{n\Qmat{\boldsymbol{\sigma}}}$, $\boldsymbol{\sigma} = \frac{1}{m\Qmat^\top\boldsymbol{\delta}}$
    \EndFor
    \State $\Tmat^{(t+1)} = \text{diag}(\boldsymbol{\delta})\Qmat\text{diag}(\boldsymbol{\sigma})$
\EndFor
\State \textbf{Return} $\Tmat$
\end{algorithmic}

\end{algorithm}


\vspace{-4mm}
\section{Proposed Method}
\label{sec:model}
\vspace{-2mm}
\subsection{Model framework overview}
\vspace{-2mm}
To capture both the topological information (network structure $\CE$) and the semantic information (text content $\CT$) in the textual network embedding, we explicitly model two types of embeddings for each node $v \in \CV$: ($i$) the topological embedding $\zv_u^t$, and ($ii$) the semantic embedding $\zv_u^{s}$. The final embedding is constructed by concatenating the topological and semantic embeddings, \ie, $\zv_u = [\zv_u^t; \zv_u^s]$. We consider the topological embedding $\zv^t$ as a static property of the node, fixed regardless of the context. On the other hand, the semantic embedding $\zv^s$ dynamically depends on the context, which is the focus of this study. 

Motivated by the work of \citep{tu2017cane}, we consider the following probabilistic objective to train the network embeddings:
\vspace{-2mm}
\beq
\ell(\Theta) = \EE_{e\sim \CE} \left\{ \ell(e;\Theta) \right\},
\vspace{-2mm}
\eeq
where $e=(u,v)$ represents sampled edges from the network and $\Theta = \{ \Zmat, \theta \}$ is the collection of model parameters. The edge loss $\ell(e;\Theta)$ is given by the cross entropy 
\beq
\ell(e_{uv};\Theta) = - w_{uv} \log p_{\Theta}(u|v), 
\label{eq:mdl_loglik}
\eeq
where $p_{\Theta}(u|v)$ denotes the conditional likelihood of observing a (weighted) link between nodes $u$ and $v$, with the latter serving as the context. More specifically,
\vspace{-2mm}
\beq
p_{\Theta}(u|v) = \langle \zv_u,  \zv_v \rangle - \log(Z),
\vspace{-2mm}
\eeq
where $\small Z = \sum_{u'\in\CV} \exp(\langle \zv_{u'}, \zv_{v} \rangle)$ is the normalizing constant and $\langle \cdot, \cdot \rangle$ is an inner product operation, to be defined momentarily. Note here we have suppressed the dependency on $\Theta$ to simplify notation. 

To capture both the topological and semantic information, along with their interactions, we propose to use the following decomposition for our inner product term: 
\vspace{-2mm}
\beq
\begin{array}{l}
\langle \zv_u, \zv_v \rangle = \underbrace{\langle \zv_u^t, \zv_v^t \rangle_{tt}}_{\text{topology}} + \underbrace{\langle \zv_u^s, \zv_v^s \rangle_{ss}}_{\text{semantic}} + \\
\hspace{5em} \underbrace{\langle \zv_u^t, \zv_v^s\rangle_{ts} + \langle \zv_u^s, \zv_v^t\rangle_{st}}_{\text{interaction}}
\end{array}
\vspace{-2mm}
\eeq
Here we use $\langle \zv_u^a, \zv_v^b \rangle_{ab}\,,a,b\in\{s,t\}$ to denote the inner product evaluation between the two feature embeddings $\zv_u^a$ and $\zv_v^b$, which can be defined by a semi-positive-definite kernel function $\kappa_{ab}(\zv_u^a,\zv_v^b)$
~\citep{alvarez2012kernels}, \eg, Euclidean kernel, Gaussian RBF, IMQ kernel, {\it etc.} Note that for $a\neq b$, $\zv_u^a$ and $\zv_v^b$ do not reside on the same feature space. As such, embeddings are first mapped to the same feature space for inner product evaluation. In this study, we use the Euclidean kernel \vspace{-2mm}
$$\langle \xv_1, \xv_2 \rangle_{\CX} = \xv_1^T \xv_2 \vspace{-2mm}$$
for inner product evaluation with $\xv_1,\xv_2 \in \CX \subseteq \BR^d$, and linear mapping 
\vspace{-2mm} $$\langle \xv, \yv \rangle_{\CX\CY} = \langle \xv, \Amat \yv\rangle_{\CX}, \text{ where } \Amat \in \BR^{d \times d'} \vspace{-2mm}$$
for feature space realignment with $\xv \in \CX \subseteq \BR^d, \yv \in \CY \subseteq \BR^{d'}$. Here $\Amat$ is a trainable parameter, and throughout this paper we omit the bias terms in linear maps to avoid notational clutter. 

Note that our solution differs from existing network-embedding models in that: ($i$) our objective is a principled likelihood loss, while prior works heuristically combine the losses of four different models ~\citep{tu2017cane}, which may fail to capture the non-trivial interactions between the fixed and dynamic embeddings; and ($ii$) we present a formal derivation of network embedding in a reproducing kernel Hilbert space. 
\vspace{-2mm}
\paragraph{Negative sampling.} Direct optimization of (\ref{eq:mdl_loglik}) requires summing over all nodes in the network, which can be computationally infeasible for large-scale networks. 
To alleviate this issue, we consider other more computationally efficient surrogate objectives. In particular, we adopt the negative sampling approach \citep{mikolov2013distributed}, which replaces the bottleneck Softmax with a more tractable approximation given by
\vspace{-2mm}
\beq
\begin{array}{l}
\log p(v|u) \approx  \log \sigma(\langle \zv_u, \zv_v \rangle) + \\
[3pt]
\hspace{5em} \sum_{j=1}^K \EE_{v_k \sim p_n}[\log \sigma(-\langle \zv_u, \zv_{v_k} \rangle)] ,
\end{array}
\vspace{-2mm}
\eeq
where $\sigma(x) = \frac{1}{1+\exp(-x)}$ is the sigmoid function, and $p_n(v)$ is a noise distribution over the nodes. 
Negative sampling can be considered as a special variant of noise contrastive estimation \citep{gutmann2010noise}, which seeks to recover the ground-truth likelihood by contrasting data samples with noise samples, thereby bypassing the need to compute the normalizing constant. As the number of noise samples $K$ goes to infinity, this approximation becomes exact\footnote{This is a non-trivial result, for completeness we provide an intuitive justification in Supplementary Material.} \citep{goldberg2014word2vec}.
Following the practice of \citet{mikolov2013distributed}, we set our noise distribution to $p_n(v) \propto d_v^\frac{3}{4}$, where $d_v$ denotes the out-degree of node $v$.

\vspace{-2mm}
\paragraph{Context matching.} We argue that a key to the context-aware network embedding is the design of an effective attention mechanism, which cross-matches the relevant content between the node's associated text and the context. Over-simplified dot-product attention limits the potential of existing textual network embedding schemes. In the following sections, we present two novel, efficient attention designs that fulfill the desiderata listed in our Introduction. Our discussion follows the setup used in CANE \citep{tu2017cane} and WANE~\citep{shen2018improved}, where the text from the interacting node is used as the context. Generalization to other forms of context is straightforward. 
\vspace{-2mm}
\subsection{Optimal-transport-based matching} 
\vspace{-2mm}
\begin{figure}[!t]
	\centering
	\includegraphics[width=.5\textwidth]{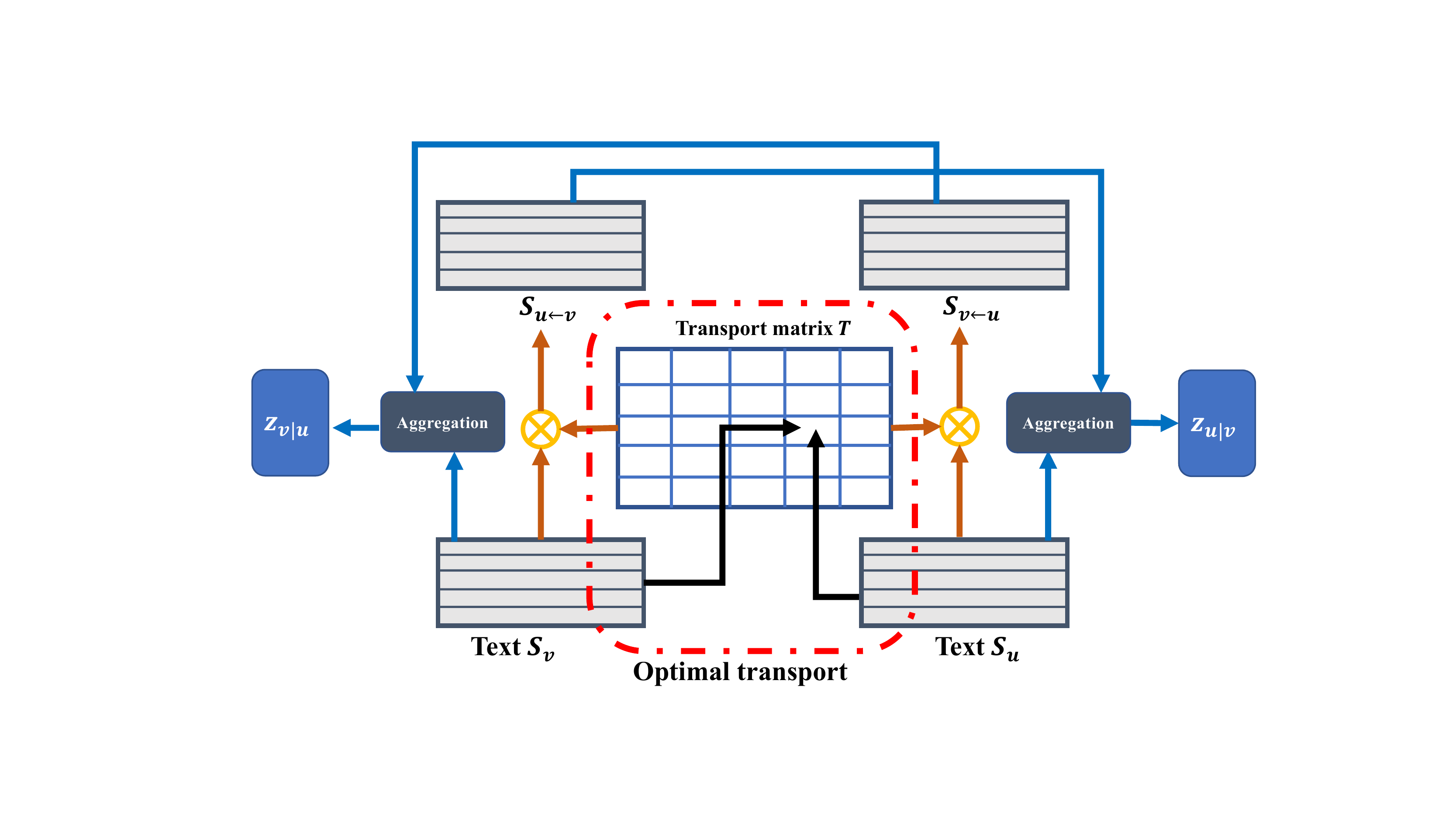}
	\caption{Schematic of the proposed mutual attention mechanism. In this setup, bag-of-words feature matchings are explicitly abstracted to infer the relationship between vertices.}
	\label{fig:mutual attention}
	\vspace{-5mm}
\end{figure}

We first consider reformulating content matching as an optimal transport problem, and then re-purpose the transport plan as our attention score to aggregate context-dependent information. More specifically, we see a node's text and context as two (discrete) distributions over the content space. Related content will be matched in the sense that they yield a higher weight in the optimal transport plan $\pi^*$. The following two properties make the optimal transport plan more appealing for use as attention score. ($i$) Sparsity: when solved exactly, $\pi^*$ is a sparse matrix with at most $(2m-1)$ non-zero elements, where $m$ is the number of contents (\citet{brualdi1991combinatorial}, $\S 8.1.3$); ($ii$) Self-normalized: row-sum and column-sum equal the respective marginal distributions. 

Implementation-wise, we first feed embedded text sequence $S_u$ and context sequence $S_v$ into our OT solver to compute the OT plan, 
\vspace{-2mm}
\beq 
\Tmat_{uv} = \text{SolveOT}(S_u, S_v)  \in \BR^{n_u \times n_v}. 
\vspace{-2mm}
\eeq
Note that here we treat pre-embedded sequence $S_u$ as $n_u$ point masses in the feature space, each with weight $1/n_u$, and similarly for $S_v$. 
Next we ``transport'' the semantic content from context $S_v$ according to the estimated OT plan with matrix multiplication
\vspace{-2mm}
\beq
S_{u\leftarrow v} = \Tmat_{uv} S_v \in \BR^{n_u \times p}\,, 
\label{eq:select}
\vspace{-2mm}
\eeq
where we have treated $S_v$ as a $\BR^{n_v \times p}$ matrix. Intuitively, this operation {\it aligns} the context with the target text sequence via averaging the context semantic embeddings with respect to the OT plan for each content element in $S_u$. To finalize the contextualized embedding, we {\it aggregate} the information from both $S_u$ and the aligned $S_{u\leftarrow v}$ with an operator $F_{\text{agg}}$,
\vspace{-2mm}
\beq 
\zv_{u|v} = F_{\text{agg}}(S_u, S_{u\leftarrow v}) \in \BR^{d \times 1}. 
\label{eq:agg}
\vspace{-2mm}
\eeq
In this case, we practice the following simple aggregation strategy: first concatenate $S_u$ and the aligned $S_{u\leftarrow v}$ along the feature dimension, and then take max-pooling along the temporal dimension to reduce the feature vector into a $2p$ vector, followed by a linear mapping to project the embedding vector to the desired dimensionality.

\vspace{-2mm}
\subsection{Attention parsing} 
\vspace{-2mm}
Direct application of attention scores based on a low-level similarity-based matching criteria (\eg, dot-product attention) can be problematic in a number of ways: ($i$) low-level attention scores can be noisy (\ie, spurious matchings), and ($ii$) similarity-matching does not allow relational inference. To better understand these points, consider the following cases. For ($i$), if the sequence embeddings used do not explicitly address the syntactic structure of the text, a relatively dense attention score matrix can be expected. For ($ii$), consider the case when the context is a query, and the matching appears as a cue in the node's text data; then the information needed is actually in the vicinity rather than the exact matching location (\eg, shifted a few steps ahead). Inspired by the work of \citet{wang2018joint}, we propose a new mechanism called {\it attention parsing} to address the aforementioned issues. 

As the name suggests, attention parsing re-calibrates the raw low-level attention scores to better integrate the information. To this end, we conceptually treat the raw attention matrix $\Tmat_{\raw}$ as a two-dimensional image and apply convolutional filters to it:
\vspace{-1mm}
\beq 
\Hmat = \text{ReLU}(\text{Conv2d}(\Tmat_{\raw}, \Wmat_F)) \in \BR^{n_u \times n_v \times c}\,, 
\vspace{-1mm}
\eeq 
where $\Wmat_F \in \BR^{h \times w \times c}$ denotes the filter banks with $h, w$ and $c$ respectively as window sizes and channel number. We can stack more convolutional layers, break sequence embedding dimensions to allow multi-group (channel) low-level attention as input, or introduce more-sophisticated model architectures (\eg, ResNet~\citep{he2016deep}, Transformer~\cite{vaswani2017attention}, {\it etc}.) to enhance our model. For now, we focus on the simplest model described above, for the sake of demonstration. 

With $\Hmat \in \BR^{n_u \times n_v \times c}$ as the high-level representation of attention, our next step is to reduce it to a weight vector to align information from the context $S_v$. We apply a max-pooling operation with respect to the context dimension, followed by a linear map to get the logits $\hv\in \BR^{n_u \times 1}$ of the weights
\beq
\hv = \text{MaxPool}(\Hmat, \text{column}) \cdot \Bmat , 
\eeq 
where $\Bmat \in \BR^{c \times 1}$ is the projection matrix. Then the parsed attention weight $\wv$ is obtained by 
\beq
\wv = \text{Softmax}(\hv)  \in \BR^{n_u \times 1}\,, 
\vspace{-2mm}
\eeq
which is used to compute the aligned context embedding
\vspace{-1mm}
\beq
\sv_{u\leftarrow v} = \wv^T S_v \in \BR^{1 \times p}. 
\vspace{-2mm}
\eeq
Note that here we compute a globally aligned context embedding vector $\sv_{u\leftarrow v}$, rather than one for each location in $S_u$ as described in the last section ($S_{u\leftarrow v}$). In the subsequent aggregation operation, $\sv_{u\leftarrow v}$ is broadcasted to all the locations in $S_u$. We call this {\it global alignment}, to distinguish it from the {\it local alignment} strategy described in the last section. Both alignment strategies have their respective merits, and in practice they can be directly combined to produce the final context-aware embedding.

\vspace{-2mm}
\section{Related Work}
\label{sec:related}
\vspace{-2mm}
\paragraph{Network embedding models.} Prior network embedding 
solutions can be broadly classified into two categories: ($i$) {\it topology embedding}, which only uses the link information; and ($ii$) {\it fused embedding}, which also exploits side information associated with the nodes. Methods from the first category focus on encoding high-order network interactions in a scalable fashion, such as LINE \citep{tang2015line}, DeepWalk \citep{perozzi2014deepwalk}.
However, models based on topological embeddings alone often ignore rich heterogeneous information associated with the vertices. Therefore, the second type of model tries to incorporate text information to improve network embeddings. 
For instance, TADW \citep{yang2015network}, CENE \cite{sun2016general}, CANE \cite{tu2017cane},
WANE~\cite{shen2018improved}, and DMTE~\cite{zhang2018diffusion}.
\vspace{-3mm}
\paragraph{Optimal Transport in NLP.} OT has found increasing application recently in NLP research. It has been successfully applied in many tasks, such as topic modeling~\citep{kusner2015word}, text generation~\citep{chen2018adversarial}, sequence-to-sequence learning~\citep{chen2019improving}, and word-embedding alignment~\citep{alvarez2018gromov}. Our model is fundamentally different from these existing OT-based NLP models in terms of how OT is used: these models all seek to minimize OT distance to match sequence distributions, while our model used the OT plan as an attention mechanism to integrate context-dependent information.
\vspace{-7mm}
\paragraph{Attention models.} Attention was originally proposed in QA systems \cite{weston2014memory} to overcome the limitations of the sequential computation associated with recurrent models~\citep{hochreiter2001gradient}. Recent developments, such as the Transformer model~\citep{vaswani2017attention}, have popularized attention as an integral part of compelling sequence models. While simple attention mechanisms can already improve model performance \cite{bahdanau2014neural, luong2015effective}, significant gains can be expected from more delicate designs~\citep{yang2016hierarchical, li2015hierarchical}. Our treatment of attention is inspired by the LEAM model~\citep{wang2018joint}, which significantly improves mutual attention in a computationally efficient way. 

\vspace{-3mm}
\section{Experiments}
\vspace{-2mm}
\subsection{Experimental setup}
\vspace{-2mm}
\paragraph{Datasets and tasks.} We consider three benchmark datasets:
($i$) {\it Cora}\footnote{\url{https://people.cs.umass.edu/\~mccallum/data.html}}, a paper citation network with text information, built by \citet{mccallum2000automating}. We prune the dataset so that it only has papers on the topic of machine learning. ($ii$) {\it Hepth}\footnote{\url{https://snap.stanford.edu/data/cit-HepTh.html}}, a paper citation network from Arxiv on high energy physics theory, with paper abstracts as text information. ($iii$) {\it Zhihu}, a Q\&A network dataset constructed by \cite{tu2017cane}, which has 10,000 active users with text descriptions and their collaboration links. Summary statistics of these three datasets are summarized in Table \ref{tab:datasets}. Pre-processing protocols from prior studies are used for data preparation \citep{shen2018improved, zhang2018diffusion, tu2017cane}.

\begin{table}[t] \small
	\begin{center}
	\resizebox{0.7\linewidth}{!}{
	\begin{tabular}{ c c c c  }
	\toprule
	 & {\it Cora} & {\it Hepth} & {\it Zhihu} \\
	\midrule
 	\#vertices & 2,227 & 1,038 & 10,000\\ 
	\#edges & 5,214 & 1,990 & 43,894\\ 
	\#avg text len & 90 & 54 & 190  \\
	\#labels & 7 & NA & NA  \\ 
 	\bottomrule   
	\end{tabular}}
	\caption{Dataset statistics. \label{tab:datasets}}
	\vspace{-3mm}
	\end{center}
	\vspace{-4mm}
\end{table}

\vspace{-1mm}
For quantitative evaluation, we tested our model on the following tasks: ($a$) {\it Link prediction}, where we deliberately mask out a portion of the edges to see if the embedding learned from the remaining edges can be used to accurately predict the missing edges. ($b$) {\it Multi-label node classification}, where we use the learned embedding to predict the labels associated with each node. Note that the label information is not used in our embedding. We also carried out ablation study to identify the gains. In addition to the quantitative results, we also visualized the embedding and the attention matrices to qualitatively verify our hypotheses.

\begin{table*}[t!]  \footnotesize 
	\centering
	\resizebox{0.8\linewidth}{!}{
	\begin{tabular}{ccccccccccc}
        \toprule[1.2pt]
        &  \multicolumn{5}{c}{Cora} & \multicolumn{5}{c}{Hepth}\\
        \hline
		\textbf{\%Training Edges} &  	\textbf{15\%} & \textbf{35\%} & 	\textbf{55\%} & \textbf{75\%} & \textbf{95\%} & \textbf{15\%} & \textbf{35\%} & \textbf{55\%} & \textbf{75\%} & \textbf{95\%}\\
		\midrule
		\textbf{MMB}      & 54.7 & 59.5 & 64.9 & 71.1 &75.9 & 54.6 &57.3 & 66.2  &73.6 &80.3 \\
		\textbf{node2vec} & 55.9 & 66.1  & 78.7 & 85.9& 88.2 & 57.1 &69.9  &84.3 & 88.4 & 89.2 \\
		\textbf{LINE}     & 55.0 & 66.4 & 77.6 & 85.6 & 89.3 & 53.7  &66.5  &78.5  &87.5 & 87.6\\ 
		\textbf{DeepWalk} & 56.0 & 70.2 & 80.1 & 85.3 & 90.3 & 55.2 &70.0& 81.3  &87.6 & 88.0  \\ 
		\midrule 
		\textbf{Naive combination} & 72.7 & 84.9 & 88.7 & 92.4 & 94.0 & 78.7 & 84.7 &88.7  &92.1 & 92.7\\ 
		\textbf{TADW} & 86.6 & 90.2 & 90.0 & 91.0& 92.7 & 87.0  &91.8& 91.1  &93.5 & 91.7 \\ 
		\textbf{CENE} & 72.1 &84.6& 89.4 & 93.9&95.5 & 86.2  &89.8 &92.3 &93.2 & 93.2\\
		\textbf{CANE} & 86.8 &92.2 &94.6 &95.6 &97.7 & 90.0  &92.0  &94.2  &95.4 & 96.3\\
		\textbf{DMTE}  & 91.3 & 93.7  & 96.0  & 97.4 & 98.8 & NA & NA & NA & NA & NA\\
		\textbf{WANE}  & 91.7 & 94.1 & 96.2 & 97.5 & 99.1 &92.3 & 95.7 & 97.5 & 97.7 & 98.7 \\
		\midrule
		\textbf{GANE-OT} & 92.0 & 95.7 & 97.3 & 98.6& 99.2 &93.4 & 97.0 & 97.9 & 98.2 & 98.8\\
		\textbf{GANE-AP} & \textbf{94.0} & \textbf{97.2} & \textbf{98.0} & \textbf{98.8} & \textbf{99.3} &\textbf{93.8} & \textbf{97.3} & \textbf{98.1} & \textbf{98.4} & \textbf{98.9} \\
		\bottomrule[1.2pt]
	\end{tabular}}
	\caption{AUC scores for link prediction on the \emph{Cora} and \emph{Hepth} dataset.}
	\label{tab:combine}
	\vspace{-2mm}
\end{table*}

\begin{table*} [!ht]  \footnotesize 
	\centering
	\resizebox{0.8\linewidth}{!}{
	\begin{tabular}{cccccccccc}
		\toprule
		\textbf{\%Training Edges} &  	\textbf{15\%} & \textbf{25\%} & 	\textbf{35\%} & \textbf{45\%} & \textbf{55\%} & \textbf{65\%} & \textbf{75\%} & \textbf{85\%} & \textbf{95\%} \\
		\midrule
		\textbf{DeepWalk}    & 56.6 & 58.1 & 60.1 & 60.0 & 61.8 & 61.9 & 63.3 & 63.7 & 67.8   \\ 
		\textbf{node2vec}        & 54.2 & 57.1 & 57.3 & 58.3 & 58.7 & 62.5 & 66.2 & 67.6 & 68.5  \\
		\textbf{LINE}            & 52.3 & 55.9 & 59.9 & 60.9 & 64.3 & 66.0 & 67.7 & 69.3 & 71.1  \\ 
		\textbf{MMB}      & 51.0 & 51.5 & 53.7 & 58.6 & 61.6 & 66.1 & 68.8 & 68.9 & 72.4  \\
		\midrule 
		\textbf{Naive combination}       & 55.1 & 56.7 & 58.9 & 62.6 & 64.4 & 68.7 & 68.9 & 69.0 & 71.5  \\ 
		\textbf{TADW}            & 52.3 & 54.2 & 55.6 & 57.3 & 60.8 & 62.4 & 65.2 & 63.8 & 69.0  \\
		\textbf{CENE} & 56.2 & 57.4 & 60.3 & 63.0 & 66.3 & 66.0 & 70.2 & 69.8 & 73.8 \\
		\textbf{CANE}  & 56.8 & 59.3 & 62.9 & 64.5 & 68.9 & 70.4 & 71.4 & 73.6 & 75.4  \\
		\textbf{DMTE} & 58.4 & 63.2 & 67.5 & 71.6 & 74.0 & 76.7 & 78.5 & 79.8 & 81.5\\
		\textbf{WANE}  & 58.7 & 63.5 & 68.3 & 71.9 & 74.9 & 77.0 & 79.7 & 80.0 & 82.6  \\
		\midrule
		\textbf{GANE-OT} & 61.6 & 66.4 & 70.8 & 73.0 & 77.3 & 80.6 & 80.4 & 81.8 & 83.2\\
		\textbf{GANE-AP} & \textbf{64.6} & \textbf{69.4} & \textbf{72.8} & \textbf{74.2} & \textbf{79.1} & \textbf{82.6} & \textbf{81.8} & \textbf{83.0} & \textbf{84.3} \\
		\bottomrule
	\end{tabular}}
	\caption{AUC scores for link prediction on the \emph{Zhihu} dataset.}
	\label{tab:zhihu}
	\vspace{-5mm}
\end{table*}

\vspace{-2mm}
\paragraph{Evaluation metrics.} For the link prediction task, we adopt the {area under the curve} (AUC) score to evaluate the performance,
AUC is employed to measure the probability that vertices in existing edges are more similar than those in the nonexistent edge.
For each training ratio, the experiment is executed $10$ times and the mean AUC scores are reported, where higher AUC indicates better performance.
For multi-label classification, we evaluate the performance with Macro-F1 scores.
The experiment for each training ratio is also executed 10 times and the average Macro-F1 scores are reported, where a higher value indicates better performance.
\vspace{-3mm}
\paragraph{Baselines.} To demonstrate the effectiveness of the proposed solutions, we evaluated our model along with the following strong baselines. ($i$) {\it Topology only embeddings}: MMB \citep{airoldi2008mixed}, DeepWalk \citep{perozzi2014deepwalk}, LINE \citep{tang2015line}, Node2vec \citep{grover2016node2vec}. ($ii$) {\it Joint embedding of topology \& text}: Naive combination, TADW \citep{yang2015network}, CENE \citep{sun2016general}, CANE \citep{tu2017cane}, WANE \citep{shen2018improved}, DMTE \citep{zhang2018diffusion}. A brief summary of these competing models is provided in the Supplementary Material (SM). 
\vspace{-3mm}
\subsection{Results}
\vspace{-2mm}
We consider two variants of our model, denoted as  GANE-OT and GANE-AP.  
GANE-OT employs the most basic OT-based attention model, specifically, global word-by-word alignment model; while GANE-AP additionally uses a one-layer convolutional neural network for the attention parsing. Detailed experimental setups are described in the SM.

\vspace{-3mm}
\paragraph{Link prediction.}

Tables \ref{tab:combine} and \ref{tab:zhihu} summarize the results from the link-prediction experiments on all three datasets, where a different ratio of edges are used for training. 
Results from models other than GANE are collected from \citet{tu2017cane}, \citet{shen2018improved} and  \citet{zhang2018diffusion}. We have also repeated these experiments on our own, and the results are consistent with the ones reported. Note that \citet{zhang2018diffusion} did not report results on DMTE. 
Both GANE variants consistently outperform competing solutions. In the low-training-sample regime our solutions lead by a large margin, and the performance gap closes as the number of training samples increases.
This indicates that our OT-based mutual attention framework can yield more informative textual representations than other methods. Note that GANE-AP delivers better results compared with GANE-OT, suggesting the attention parsing mechanism can further improve the low-level mutual attention matrix. More results on {\it Cora} and {\it Hepth} are provided in the SM.

\vspace{-3mm}
\paragraph{Multi-label Node Classification.} 
To further evaluate the effectiveness of our model, we consider multi-label vertex classification. Following the setup described in \citep{tu2017cane}, we first computed all context-aware embeddings. Then we averaged over each node's context-aware embeddings with all other connected nodes, to obtain a global embedding for each node, \ie,  $\zv_u = \frac{1}{d_u}\sum_v \zv_{u|v}$, where $d_u$ denotes the degree of node $u$. 
A linear SVM is employed, instead of a sophisticated deep classifier, to predict the label attribute of a node. We randomly sample a portion of labeled vertices with embeddings (${10\%, 30\%, 50\%, 70\%}$) to train the classifier, using the rest of the nodes to evaluate prediction accuracy.
We compare our results with those from other state-of-the-art models in Table \ref{tab:cora_cls}. The GANE models delivered better results compared with their counterparts, lending strong evidence that the OT attention and attention parsing mechanism promise to capture more meaningful representations.
	

\begin{table}[t!]  \footnotesize 
	\centering
	
	\begin{tabular}{ccccc}
		\toprule
		\textbf{\%training labels} & \textbf{10\%} & \textbf{30\%} & \textbf{50\%} & \textbf{70\%} \\
		\midrule
        \textbf{LINE}  &  53.9 & 56.7 & 58.8 & 60.1 \\
        \textbf{TADW}  &  71.0 & 71.4 & 75.9 & 77.2 \\
        \textbf{CANE}  &  81.6 & 82.8 & 85.2 & 86.3 \\
        \textbf{DMTE}  &  81.8 & 83.9 & 86.3 & 87.9 \\
        \textbf{WANE}  &  81.9 & 83.9 & 86.4 & 88.1 \\
        \midrule
        \textbf{GANE-OT} & 82.0 & 84.1 & 86.6 & 88.3 \\
        \textbf{GANE-AP} & \textbf{82.3} & \textbf{84.2} & \textbf{86.7} & \textbf{88.5} \\
		\bottomrule
	\end{tabular}
	\caption{Test Macro-F1 scores for multi-label node classification on {\it Cora}.}
	\label{tab:cora_cls}
	\vspace{-2mm}
\end{table}

\begin{figure}[t!]
\small
    \centering
    \includegraphics[width=.80\columnwidth]{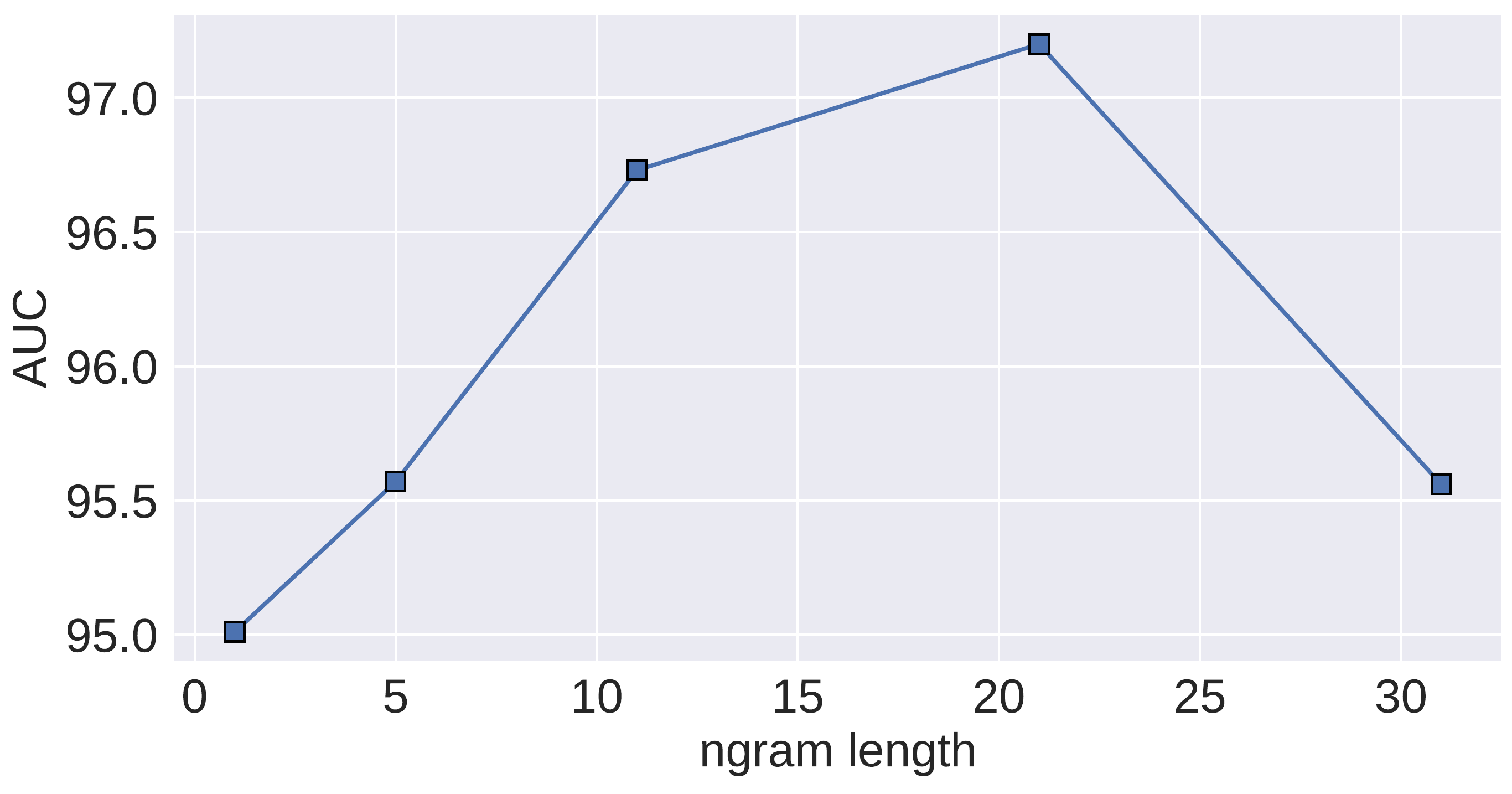}
    \caption{\small $n$-gram length {\it VS} AUC on Cora.}
    \vspace{-3mm}
    \label{fig:length vs auc plot}
    \vspace{-2mm}
\end{figure}

\vspace{-2mm}
\paragraph{Ablation study.} \label{subsec: ablation}
We further explore the effect of $n$-gram length in our model (\ie, the filter size for the covolutional layers used by the attention parsing module). In Figure \ref{fig:length vs auc plot} we plot the AUC scores for link prediction on the Cora dataset against varying $n$-gram length. The performance peaked around length $20$, then starts to drop, indicating a moderate attention span is more preferable. 
Similar results are observed on other datasets (results not shown). Experimental details on the ablation study can be found in the SM.

\begin{figure}[t!]
    \centering
    \includegraphics[width=5.5cm, height=3.7cm, scale=0.6]{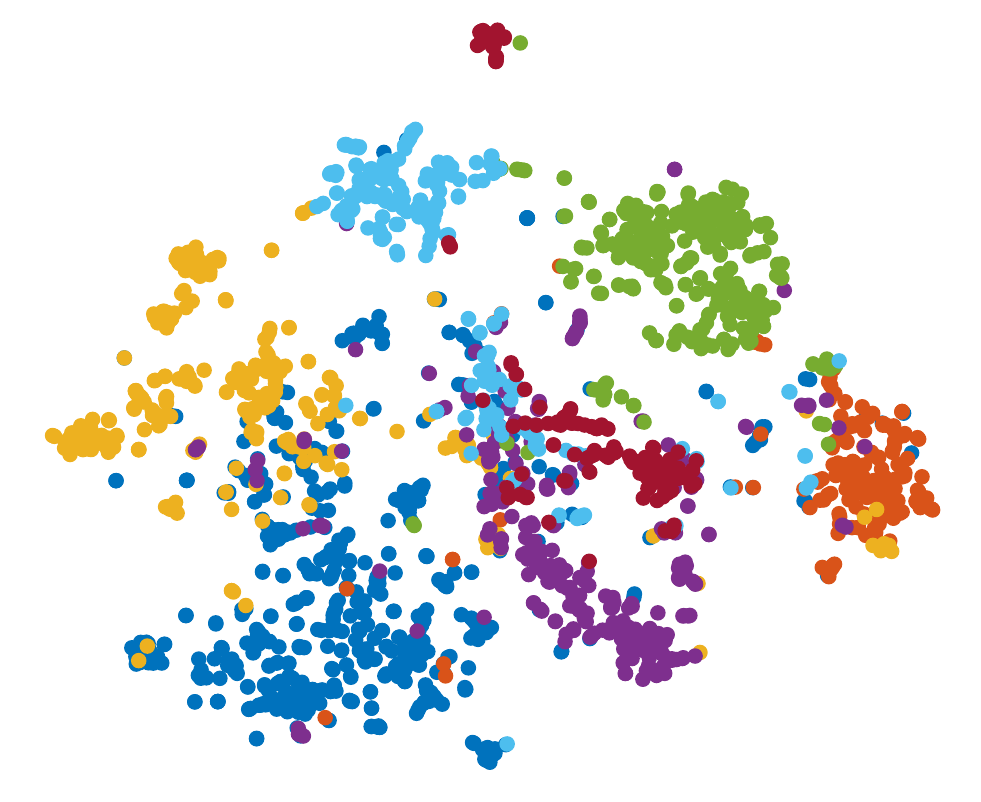}
    \vspace{-3mm}
    \caption{\small t-SNE visualization on Cora dataset.}
    \label{fig:t-sne}
    \vspace{-3mm}
\end{figure}

\vspace{-3mm}
\subsection{Qualitative Analysis}
\vspace{-1mm}
\paragraph{Embedding visualization.} 
We employed t-SNE~\citep{maaten2008visualizing} to project the network embeddings for the {\it Cora} dataset in a two-dimensional space using GANE-OT, with each node color coded according to its label. As shown in Figure \ref{fig:t-sne}, papers clustered together belong to the same category, with the clusters well-separated from each other in the network embedding space. 
Note that our network embeddings are trained without any label information. Together with the label classification results, this implies our model is capable of extracting meaningful information from both context and network topological.

\begin{figure}[t!]
    \centering
    \includegraphics[width=.9\columnwidth]{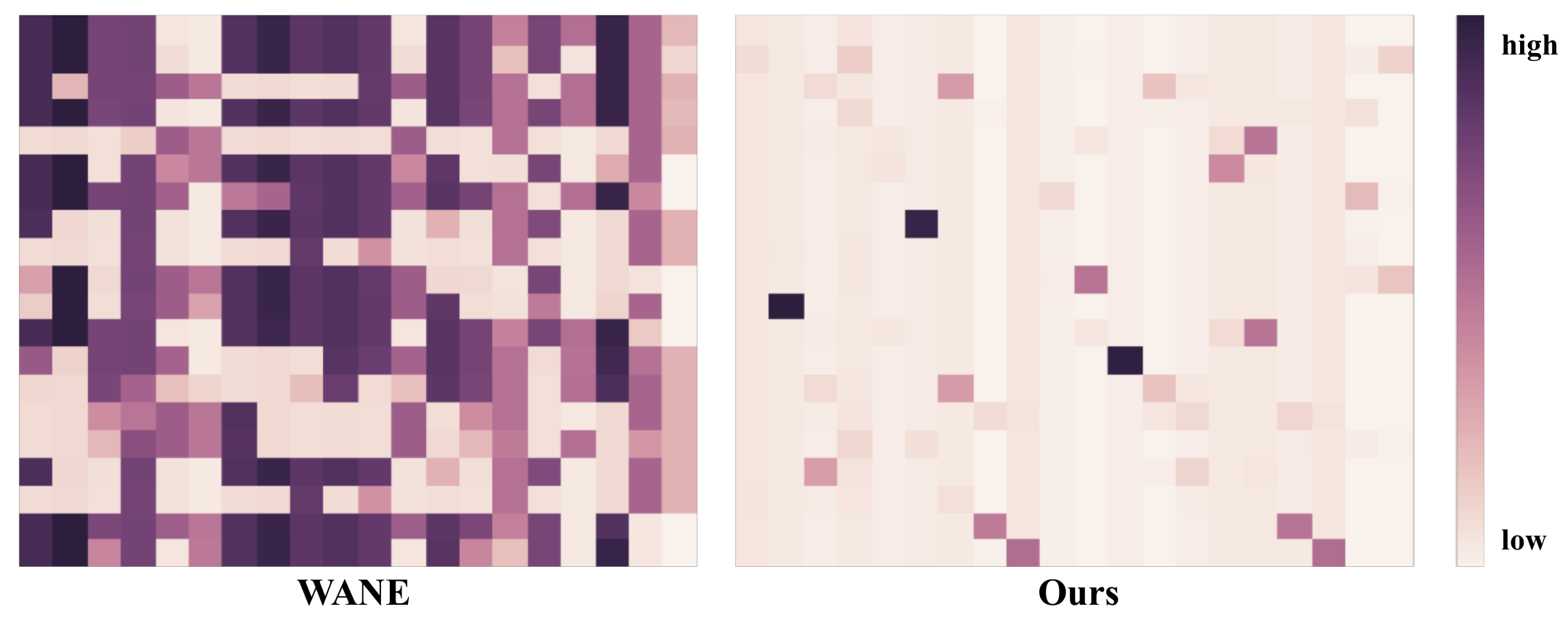}
    \vspace{-3mm}
    \caption{Mutual attention between two nodes in Cora. Left: WANE attention. Right: OT attention (ours).}
    \label{fig:att_compare}
    \vspace{-3mm}
\end{figure}

\vspace{-3mm}
\paragraph{Attention matrix comparison.} To verify that our OT-based attention mechanism indeed produces sparse attention scores, we visualized the OT attention matrices and compared them with those simarlity-based attention matrices (\eg, WANE). Figure \ref{fig:att_compare} plots one typical example. Our OT solver returns a sparse attention matrix, while dot-product-based WANE attention is effectively dense. This underscores the effectiveness of OT-based attention in terms of noise suppression.

\vspace{-3mm}
\section{Conclusion}
\vspace{-3mm}
We have proposed a novel and principled mutual-attention framework based on optimal transport (OT). Compared with existing solutions, the attention mechanisms employed by our GANE model enjoys the following benefits: ({\it i}) it is naturally sparse and self-normalized, ({\it ii}) it is a global sequence matching scheme, and ({\it iii}) it can capture long-term interactions between two sentences. 
These claims are supported by experimental evidence from link prediction and multi-label vertex classification. Looking forward, our attention mechanism can also be applied to tasks such as relational networks~\citep{santoro2017simple}, natural language inference~\citep{maccartney2009natural}, and QA systems~\citep{zhou2015learning}.

\section*{Acknowledgments}
This research was supported in part by DARPA, DOE, NIH,
ONR and NSF.
\bibliography{acl2019}
\bibliographystyle{acl_natbib}

\appendix
\label{sec:appendix}
\section{Appendix}
\subsection{Competing models} \label{sm:models}

\subsubsection{Topology only embeddings}

\noindent {\it Mixed Membership Stochastic Blockmodel} (MMB) \citep{airoldi2008mixed}: a graphical model for relational data, each node randomly select a different "topic" when forming an edge.

\noindent DeepWalk \citep{perozzi2014deepwalk}: executes truncated random walks on the graph, and by treating nodes as tokens and random walks as natural language sequences, the node embedding are obtained using the SkipGram model \citep{mikolov2013distributed}. 

\noindent Node2vec \citep{grover2016node2vec}: a variant of DeepWalk by executing biased random walks to explore the neighborhood (\eg, Breadth-first or Depth-first sampling). 

\noindent {\it Large-scale Information Network Embedding} (LINE) \citep{tang2015line}: scalable network embedding scheme via maximizing the joint and conditional likelihoods. 

\subsection{Joint embedding of topology \& text}

\noindent Naive combination \citep{tu2017cane}: direct combination of the structure embedding and text embedding that best predicts edges. 

\noindent {\it Text-Associated DeepWalk} (TADW) \citep{yang2015network}: reformulating embedding as a matrix factorization problem, and fused text-embedding into the solution. 

\noindent {\it Content-Enhanced Network Embedding} (CENE) \citep{sun2016general}: treats texts as a special kind of nodes. 

\noindent {\it Context-Aware Network Embedding} (CANE) \citep{tu2017cane}: decompose the embedding into context-free and context-dependent part, use mutual attention to address the context-dependent embedding. 

\noindent {\it Word-Alignment-based Network Embedding} (WANE) \citep{shen2018improved}: Using fine-grained alignment to improve context-aware embedding. 

\noindent {\it Diffusion Maps for Textual network Embedding} (DMTE) \citep{zhang2018diffusion}: using truncated diffusion maps to improve the context-free part embedding in CANE.

\section{Complete Link prediction results on Cora and Hepth}
\label{sm:complete}
The complete results for Cora and Hepth are listed in Tables \ref{tab:cora} and \ref{tab:hepth}. Results from models other than GANE are collected from \citep{tu2017cane,shen2018improved, zhang2018diffusion}. We have also repeated these experiments on our own, the results are consistent with the ones reported. Note that DMTE did not report results on {\it Hepth} \citep{zhang2018diffusion} . 
\begin{table*}[!ht]   \footnotesize 
	\centering
	
	\begin{tabular}{|c||c|c|c|c|c|c|c|c|c|}
        \hline
		\textbf{\%Training Edges} &  	\textbf{15\%} & \textbf{25\%} & 	\textbf{35\%} & \textbf{45\%} & \textbf{55\%} & \textbf{65\%} & \textbf{75\%} & \textbf{85\%} & \textbf{95\%} \\
		\hline
		\textbf{MMB}      & 54.7 & 57.1 & 59.5 & 61.9 & 64.9 & 67.8 & 71.1 & 72.6 & 75.9  \\
		\textbf{node2vec}        & 55.9 & 62.4 & 66.1 & 75.0 & 78.7 & 81.6 & 85.9 & 87.3 & 88.2   \\
		\textbf{LINE}           & 55.0 & 58.6 & 66.4 & 73.0 & 77.6 & 82.8 & 85.6& 88.4& 89.3  \\ 
		\textbf{DeepWalk}    & 56.0 & 63.0 & 70.2 & 75.5 & 80.1 & 85.2 & 85.3 & 87.8 & 90.3   \\ 
		\hline 
		\textbf{Naive combination}       & 72.7 & 82.0 & 84.9 & 87.0 & 88.7 & 91.9 & 92.4 & 93.9 & 94.0  \\ 
		\textbf{TADW}            & 86.6 & 88.2 & 90.2 & 90.8 & 90.0 & 93.0 & 91.0 & 93.4 & 92.7  \\ 
		\textbf{CENE} & 72.1 & 86.5 &84.6& 88.1& 89.4 &89.2& 93.9 &95.0 &95.9  \\
		\textbf{CANE}  & 86.8& 91.5 &92.2 &93.9 &94.6 &94.9 &95.6 &96.6& 97.7  \\
		\textbf{DMTE}  & 91.3 & 93.1 & 93.7 & 95.0 & 96.0 & 97.1 & 97.4 & 98.2 & 98.8 \\
		\textbf{WANE}  & 91.7 & 93.3 & 94.1 & 95.7 & 96.2 & 96.9 & 97.5 & 98.2 & 99.1  \\
		\hline
		\textbf{GANE-OT} & 92.0 & 94.4 & 95.7 & 96.6 & 97.3 & 97.6 & 98.6 & 98.8 & 99.2\\
		\textbf{GANE-AP} & \textbf{94.0} & \textbf{96.4} & \textbf{97.2} & \textbf{97.4} & \textbf{98.0} & \textbf{98.2} & \textbf{98.8} & \textbf{99.1} & \textbf{99.3} \\
		\hline
	\end{tabular}
	\caption{AUC scores for link prediction on the \emph{Cora} dataset.}
	\label{tab:cora}
\end{table*}
\begin{table*}[!ht]  \footnotesize 
	\centering
	\begin{tabular}{|c||c|c|c|c|c|c|c|c|c|}
        \hline
		\textbf{\%Training Edges} &  	\textbf{15\%} & \textbf{25\%} & 	\textbf{35\%} & \textbf{45\%} & \textbf{55\%} & \textbf{65\%} & \textbf{75\%} & \textbf{85\%} & \textbf{95\%} \\
		\hline
		\textbf{MMB}      &54.6& 57.9 &57.3 &61.6& 66.2 &68.4 &73.6 &76.0 &80.3  \\
		\textbf{DeepWalk}    & 55.2& 66.0 &70.0& 75.7& 81.3 &83.3 &87.6 &88.9 &88.0   \\ 
		\textbf{LINE}            & 53.7 &60.4 &66.5 &73.9 &78.5 &83.8 &87.5 &87.7 &87.6  \\ 
		\textbf{node2vec}        & 57.1 &63.6 &69.9 &76.2 &84.3 &87.3& 88.4 &89.2 &89.2  \\
		\hline 
		\textbf{Naive combination}       & 78.7 &82.1& 84.7& 88.7 &88.7 &91.8 &92.1 &92.0 &92.7  \\ 
		\textbf{TADW}            & 87.0 &89.5 &91.8& 90.8 &91.1 &92.6 &93.5 &91.9 &91.7  \\
		\textbf{CENE} & 86.2 &84.6 &89.8 &91.2 &92.3 &91.8 &93.2 &92.9 &93.2  \\
		\textbf{CANE}  & 90.0 & 91.2 &92.0 &93.0 &94.2 &94.6 &95.4 &95.7 &96.3  \\
		\hline
		\textbf{WANE-\emph{ww}}  & 92.3 & 94.1 & 95.7 & 96.7 & 97.5 & 97.5 & 97.7 & 98.2 & 98.7  \\
		\textbf{DMTE} & - & - & - & - & - & - & - & - & - \\
		\hline
		\textbf{GANE-OT} & 93.4 & 96.2 & 97.0 & 97.7 & 97.9 & 98.0 & 98.2 & 98.6 & 98.8 \\
		\textbf{GANE-AP} & {\bf 93.8} & {\bf 96.4} & {\bf 97.3} & {\bf 97.9} & {\bf 98.1} & {\bf 98.2} & {\bf 98.4} & {\bf 98.7} & {\bf 98.9} \\
		\hline
	\end{tabular}
	\caption{AUC scores for link prediction on the \emph{Hepth} dataset.}
	\label{tab:hepth}
	\vspace{-2mm}
\end{table*}

\subsection{Negative sampling approximation}
\label{sm:nce}
In this section we provide a quick justification for the negative sampling approximation. To this end, we first briefly review {\it noise contrastive estimation} (NCE) and how it connects to maximal likelihood estimation, then we establish the link to negative sampling. Interested readers are referred to \citet{ruder2016on} for a more thorough discussion on this topic. 

\paragraph{Noise contrastive estimation.} NCE seeks to learn the parameters of a likelihood model $p_{\Theta}(u|v)$ by optimizing the following discriminative objective: 
\begin{align}
J(\Theta) = 
&\sum_{u_i\sim p_d}[\log p_{\Theta}(y=1|u_i,v) - \nonumber \\&K \EE_{\tilde{u}'\sim p_n} [\log p_{\Theta}(y=0|\tilde{u},v)]], 
\end{align}
where $y$ is the label of whether $u$ comes from the data distribution $p_d$ or the tractable noise distribution $p_n$, and $v$ is the context. Using the Monte Carlo estimator for the second term gives us 
\begin{align}
\hat{J}(\Theta) = &\sum_{u_i\sim p_d}[\log p_{\Theta}(y=1|u_i,v) -\nonumber\\& \sum_{k=1}^K [\log p_{\Theta}(y=0|\tilde{u}_k,v)]], \tilde{u}_k \stackrel{iid}{\sim} p_n. 
\end{align}
Since the goal of $J(\Theta)$ is to predict the label of a sample from a mixture distribution with $1/(K+1)$ from $p_d$ and $K/(K+1)$ from $p_n$, plugging the model likelihood and noise likelihood into the label likelihood gives us
\begin{align}
& p_{\Theta}(y=1;u,v) =  \frac{p_{\Theta}(u|v)}{p_{\Theta}(u|v) + K p_n(u|v)}, \,\nonumber\\ &p_{\Theta}(y=0;u,v) = \frac{K p_n(u|v)}{p_{\Theta}(u|v) + K p_n(u|v)}. 
\end{align}
Recall $p_{\Theta}(u|v)$ takes the following softmax form
\beq
p_{\Theta}(u|v) = \frac{\exp(\langle u, v \rangle)}{Z_v(\Theta)}, \, Z_v(\Theta) = \sum_{u'} \exp(\langle u', v \rangle). 
\eeq
NCE treats $Z_v$ as an learnable parameter and optimized along with $\Theta$. One key observation is that, in practice, one can safely clamp $Z_v$ to $1$, and the NCE learned model ($p_{\hat{\Theta}}$) will {\it self-normalize} in the sense that $Z_v(\hat{\Theta}) \approx 1$. As such, one can simply plug $p_{\Theta}(u|v) = \exp(\langle u, v \rangle))$ into the above objective. Another key result is that, as $K\rightarrow \infty$, the gradient of NCE objective recovers the gradient of softmax objective $\log p_{\Theta}(u|v)$ \citep{dyer2014notes}.  

\paragraph{Negative sampling as NCE.} If we set $K=\#(\CV)$ and let $p_n(u|v)$ be the uniform distribution on $\CV$, we have 
\beq
p_{\Theta}(y=1|u,v) = \sigma(\langle u, v \rangle), 
\eeq
where $\sigma(z)$ is the sigmoid function. Plugging this back to the $\hat{J}(\Theta)$ covers the negative sampling objective Eqn (6) used in the paper. Combined with the discussion above, we know Eqn (6) provides a valid approximation to the $\log$-likelihood in terms of the gradient directions, when $K$ is sufficiently large. In this study, we use $K=1$ negative sample for computational efficiency. Using more samples did not significantly improve our results (data not shown). 

\subsection{Experiment Setup} \label{sm:exp_setup}
We use the same codebase from CANE~\citep{tu2017cane}\footnote{\url{https://github.com/thunlp/CANE}}. The implementation is based on TensorFlow, all experiments are exectuted on a single NVIDIA TITAN X GPU. We set embedding dimension to $d=100$ for all our experiments. 
To conduct a fair comparison with the baseline models, we follow the experiment setup from \citet{shen2018improved}. For all experiments, we set word embedding dimension as 100 trained from scratch. We train the model with Adam optimizer and set learning rate $1e-3$. For GANE-AP model, we use best filte  size $1 \times 21 \times 1$ for convolution from our abalation study.

\subsection{Ablation study setup} \label{sm:ablation}
To test how the $n$-gram length affect our GANE-AP model performance, we re-run our model with different choices of $n$-gram length, namely, the window size in convolutional layer. Each experiment is repeated for $10$ times and we report the averaged results to eliminate statistical fluctuations.

\end{document}